\documentclass[english]{article}
\usepackage{geometry}
\usepackage{babel}
\usepackage{amsmath}
\usepackage{amssymb}
\usepackage{graphicx}
\usepackage{esint}
\usepackage[unicode=true,pdfusetitle,
 bookmarks=true,bookmarksnumbered=false,bookmarksopen=false,
 breaklinks=false,pdfborder={0 0 1},backref=false,colorlinks=false]
 {hyperref}

\makeatletter

\providecommand{\tabularnewline}{\\}

\usepackage{amsmath,amstext,amsfonts,amsxtra,amssymb,latexsym,bm}
\usepackage{microtype}

\makeatother

\begin{document}

\title{Choice by Elimination via Deep Neural Networks}

\author{Truyen Tran, Dinh Phung and Svetha Venkatesh\\
Centre for Pattern Recognition and Data Analytics\\
Deakin University, Geelong, Australia\\
\emph{\{truyen.tran,dinh.phung,svetha.venkatesh\}@deakin.edu.au}}
\maketitle
\begin{abstract}
We introduce \emph{Neural Choice by Elimination}, a new framework
that integrates deep neural networks into probabilistic sequential
choice models for learning to rank. Given a set of items to chose
from, the elimination strategy starts with the whole item set and
iteratively eliminates the least worthy item in the remaining subset.
We prove that the choice by elimination is equivalent to marginalizing
out the random Gompertz latent utilities. Coupled with the choice
model is the recently introduced Neural Highway Networks for approximating
arbitrarily complex rank functions. We evaluate the proposed framework
on a large-scale public dataset with over $425K$ items, drawn from
the Yahoo! learning to rank challenge. It is demonstrated that the
proposed method is competitive against state-of-the-art learning to
rank methods.
\end{abstract}
\global\long\def\wb{\mathit{\boldsymbol{w}}}
\global\long\def\bb{\mathit{\boldsymbol{b}}}
\global\long\def\zb{\mathit{\boldsymbol{z}}}
\global\long\def\xb{\mathit{\boldsymbol{x}}}
\global\long\def\yb{\mathit{\boldsymbol{y}}}
\global\long\def\ob{\mathit{\boldsymbol{o}}}
\global\long\def\x{\mathit{\mathbf{x}}}
\global\long\def\y{\mathit{\mathbf{y}}}
\global\long\def\z{\mathit{\mathbf{z}}}
\global\long\def\o{\mathit{\mathbf{o}}}
\global\long\def\w{\mathit{\mathbf{w}}}
\global\long\def\ub{\mathit{\boldsymbol{u}}}
\global\long\def\perm{\mathit{\boldsymbol{\pi}}}
\global\long\def\pib{\mathit{\boldsymbol{\pi}}}
\global\long\def\param{\mathit{\boldsymbol{\theta}}}
\global\long\def\Real{\mathit{\mathbb{R}}}
\global\long\def\Risk{\mathit{\mathcal{R}}}
\global\long\def\thetab{\boldsymbol{\theta}}

\section{Introduction}

People often rank options when making choice. Ranking is central in
many social and individual contexts, ranging from election \cite{Gormley-Murphy08},
sports \cite{henery1981permutation}, information retrieval \cite{liu2011learning},
question answering \cite{agarwal2012learning}, to recommender systems
\cite{Truyen:2011a}. We focus on a setting known as learning to rank
(L2R) in which the system learns to choose and rank items (e.g. a
set of documents, potential answers, or shopping items) in response
to a query (e.g., keywords, a question, or an user). 

Two main elements of a L2R system are rank model and rank function.
One of the most promising rank models is listwise \cite{liu2011learning}
where all items responding to the query are considered simultaneously.
Most existing work in L2R focuses on designing listwise rank losses
rather than formal models of \emph{choice}, a crucial aspect of building
preference-aware applications. One of a few exceptions is the Plackett-Luce
model which originates from Luce's axioms of choice \cite{plackett1975analysis}
and is later used in the context of L2R under the name ListMLE \cite{xia2008listwise_FULL}.
The Plackett-Luce model offers a natural interpretation of making
sequential choices. First, the probability of choosing an item is
proportional to its worth. Second, once the most probable item is
chosen, the next item will be picked from the remaining items in the
same fashion.

\begin{figure}[t]
\begin{centering}
\includegraphics[width=0.6\textwidth]{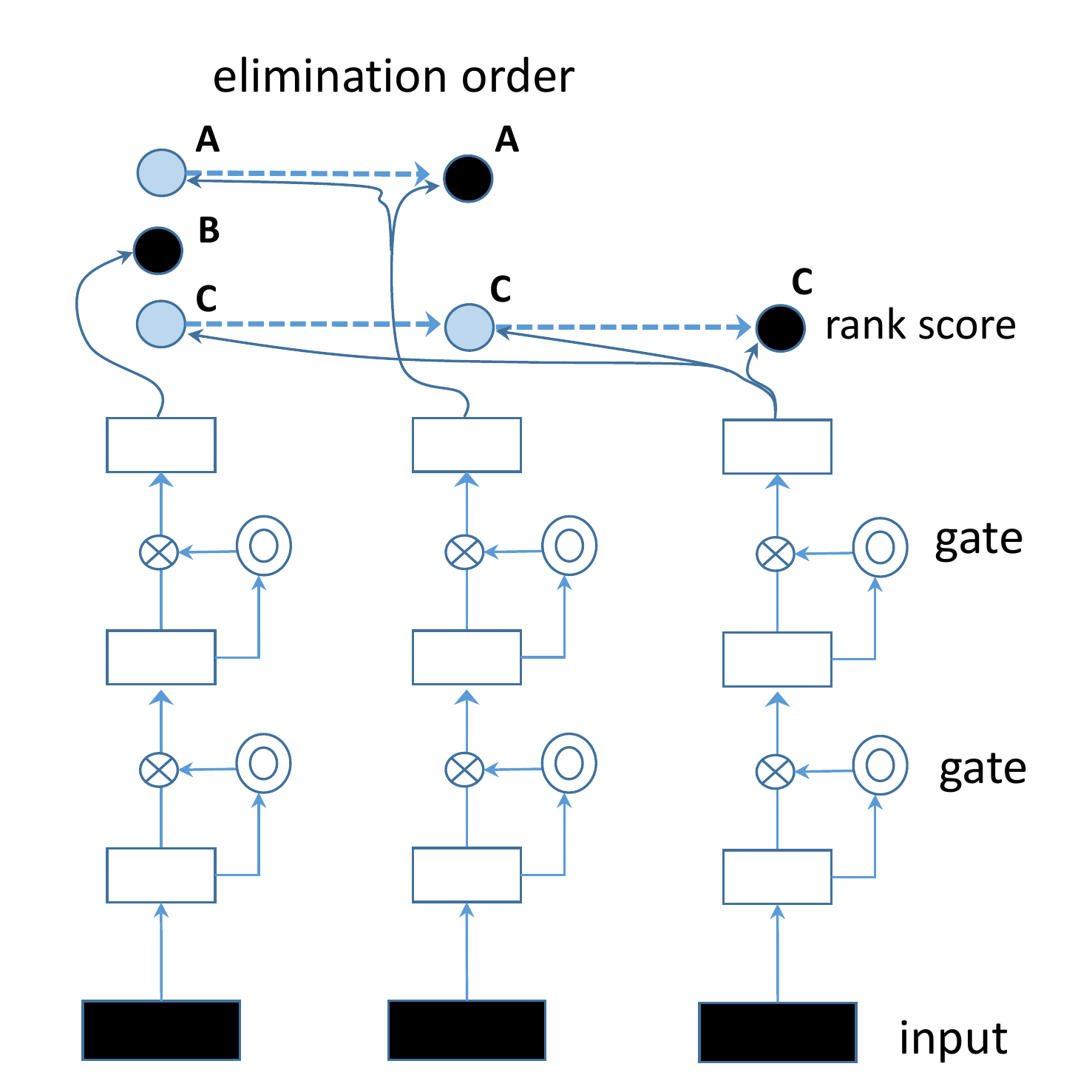}
\par\end{centering}

\protect\caption{\emph{Neural choice by elimination} with 4-layer highway networks\emph{
}for ranking three items (A,B,C). Empty boxes represent hidden layers
that share the same parameters (hence, recurrent). Double circles
are gate that controls information flow. Dark filled circles represent
a chosen item at each step -- here the elimination order is (B,A,C).
\label{fig:highway-of-rank} }
\end{figure}

However, the Plackett-Luce model suffers from two drawbacks. First,
the model spends effort to separate items down the rank list, while
only the first few items are usually important in practice. Thus the
effort in making the right ordering should be spent on more important
items. Second, the Plackett-Luce is inadequate in explaining many
competitive situations, e.g., sport tournaments and buying preferences,
where the ranking process is reversed -- worst items are eliminated
first \cite{tversky1972elimination}. 

Addressing these drawbacks, we introduce a probabilistic sequential
rank model termed \emph{choice by elimination}. At each step, we remove
one item and repeat until no item is left. The rank of items is then
the reverse of the elimination order. This elimination process has
an important property: Near the end of the process, only best items
compete against the others. This is unlike the selection process in
Plackett-Luce, where the best items are contrasted against all other
alternatives. We may face difficulty in separating items of similarly
high quality but can ignore irrelevant items effortlessly. The elimination
model thus reflects more effort in ranking worthy items. 

Once the ranking model has been specified, the next step in L2R is
to design a rank function $f(\xb)$ of query-specific item attributes
$\xb$ \cite{liu2011learning}. We leverage the newly introduced \emph{highway
networks} \cite{srivastava2015training} as a rank function approximator.
Highway networks are a compact deep neural network architecture that
enables passing information and gradient through hundreds of hidden
layers between input features and the function output. The highway
networks coupled with the proposed elimination model constitute a
new framework termed \emph{Neural Choice by Elimination (NCE)} illustrated
in Fig.~\ref{fig:highway-of-rank}.

The framework is an alternative to the current state-of-the-arts in
L2R which involve tree ensembles \cite{Friedman-ann-stat01,breiman96bagging,breiman2001random}
trained with hand-crafted metric-aware losses \cite{burges2011learning,li2007mcrank}.
Unlike the tree ensembles where typically hundreds of trees are maintained,
highway networks can be trained with \emph{dropouts} \cite{srivastava2014dropout}
to produce an implicit ensemble with only one thin network. Hence
we aim to establish that \emph{deep neural networks are competitive
in L2R}. While shallow neural networks have been used in ranking before
\cite{burges2005learning_FULL}, they were outperformed by tree ensembles
\cite{li2007mcrank}. Deep neural nets are compact and more powerful
\cite{bengio2009learning}, but they have not been measured against
tree-based ensembles for generic L2R problems. We empirically demonstrate
the effectiveness of the proposed ideas on a large-scale public dataset
from Yahoo! L2R challenge with totally $18.4$ thousands queries and
$425$ thousands documents. 

To summarize, our paper makes the following contributions: (i) introducing
a new neural sequential choice model for learning to rank; and (ii)
establishing that deep nets are scalable and competitive as rank function
approximator in large-scale settings.

\section{Background}

\subsection{Related Work}

The elimination process has been found in multiple competitive situations
such as multiple round contests and buying decisions \cite{fu2014reverse}.
\emph{Choice by elimination} of distractors has been long studied
in the psychological literature, since the pioneer work of Tversky
\cite{tversky1972elimination}. These backward elimination models
may offer better explanation than the forward selection when eliminating
aspects are available \cite{tversky1972elimination}. However, existing
studies are mostly on selecting a single best choice. Multiple sequential
eliminations are much less studied \cite{azari2012random}. Second,
most prior work has been evaluated on a handful of items with several
attributes, whereas we consider hundreds of thousands of items with
thousands of attributes. Third, the cross-field connection with data
mining has not been made. The link between choice models and Random
Utility Theory has been well-studied since Thurstone in the 1920s,
and is still an active topic \cite{azari2012random,truyen_phung_venkatesh_icml13,Truyen:2012b}.
Deep neural networks for L2R have been studied in the last two years
\cite{huang2013learning,deng2013deep,song2014adapting,dong2014rankcnn,severyn2015learning}.
Our work contributes a formal reasoning of human choices together
with a newly introduced highway networks which are validated on large-scale
public datasets against state-of-the-art methods.

\subsection{Plackett-Luce}

We now review Plackett-Luce model \cite{plackett1975analysis}, a
forward selection method in learning to rank, also known in the L2R
literature as ListMLE \cite{xia2008listwise_FULL}. Given a query
and a set of response items $\mathcal{I}=\left(1,2,..,N\right)$,
the rank choice is an ordering of items $\pib=(\pi_{1},\pi_{2},...,\pi_{N})$,
where $\pi_{i}$ is the index of item at rank $i$. For simplicity,
assume that each item $\pi_{i}$ is associated with a set of attributes,
denoted as $\xb_{\pi_{i}}\in\mathbb{R}^{p}$. A rank function $f(\xb_{\pi_{i}})$
is defined on $\pi_{i}$ and is independent of other items. We aim
to characterize the rank permutation model $P(\perm)$. 

Let us start from the classic probabilistic theory that any joint
distribution of $N$ variables can be factorized according to the
chain-rule as follows

\begin{equation}
P(\pib)=P(\pi_{1})\prod_{i=2}^{N}P(\pi_{i}\mid\boldsymbol{\pi}_{1:i-1})\label{eq:PL-factorisation}
\end{equation}
where $\boldsymbol{\pi}_{1:i-1}$ is a shorthand for $(\pi_{1},\pi_{2},..,\pi_{i-1})$,
and $P(\pi_{i}\mid\boldsymbol{\pi}_{1:i-1})$ is the probability that
item $\pi_{i}$ has rank $i$ given all existing higher ranks $1,...,i-1$.
The factorization can be interpreted as follows: choose the first
item in the list with probability of $P(\pi_{1})$, and choose the
second item from the remaining items with probability of $P(\pi_{2}|\pi_{1})$,
and so on. Luce's \emph{axioms of choice} assert that an item is chosen
with probability proportional to its \emph{worth}. This translates
to the following choice model:
\[
P(\pi_{i}\mid\boldsymbol{\pi}_{1:i-1})=\frac{\exp\left(f(\xb_{\pi_{i}})\right)}{\sum_{j=i}^{N}\exp\left(f(\xb_{\pi_{j}})\right)}
\]
Learning using maximizing likelihood minimizes the log-loss: 
\begin{align}
\ell_{1}(\pib) & =\sum_{i=1}^{N-1}\left(-f(\xb_{\pi_{i}})+\log\sum_{j=i}^{N}\exp\left(f(\xb_{\pi_{j}})\right)\right)\label{eq:PL-loss}
\end{align}

\section{Choice by Elimination \label{sec:High-order-Markov-chains}}

We note that the factorization in Eq.~(\ref{eq:PL-factorisation})
is not unique. If we permute the indices of items, the factorization
still holds. Here we derive a \emph{reverse} Plackett-Luce model as
follows 
\begin{equation}
P(\pib)=Q(\pi_{N})\prod_{i=1}^{N-1}Q(\pi_{i}\mid\boldsymbol{\pi}_{i+1:N})\label{eq:PL-reverse}
\end{equation}
where $Q(\pi_{i}\mid\boldsymbol{\pi}_{i+1:N})$ is the probability
that item $\pi_{i}$ receives rank $i$ given all existing lower ranks
$i+1,i+2,...,N$

Since $\pi_{N}$ is the most irrelevant item in the list, $Q(\pi_{N})$
can be considered as the probability of \emph{eliminating} the item.
Thus the entire process is backward elimination: The next irrelevant
item $\pi_{k}$ is eliminated, given that more extraneous items ($\pib_{i>k)}$)
have already been eliminated. It is reasonable to assume that \emph{the
probability of an item being eliminated is inversely proportional
to its worth.} This suggests the following specification

\begin{eqnarray}
Q(\pi_{i}|\boldsymbol{\pi}_{i+1:N}) & = & \frac{\exp\left(-f(\xb_{\pi_{i}})\right)}{\sum_{j=1}^{i}\exp\left(-f(\xb_{\pi_{j}})\right)}.\label{eq:backward-local-prob}
\end{eqnarray}
Note that, due to specific choices of conditional distributions, distributions
in Eqs.~(\ref{eq:PL-factorisation},\ref{eq:PL-reverse}) are generally
not the same. With this model, the log-loss has the following form:
\begin{align}
\ell_{2}(\perm) & =\sum_{i=1}^{N}\left(f(\xb_{\pi_{i}})+\log\sum_{j=1}^{i}\exp\left(-f(\xb_{\pi_{j}})\right)\right)\label{eq:rev-PL-loss}
\end{align}

\subsection{Derivation using Random Utility Theory.}

Random Utility Theory \cite{azari2012random,thurstone1927law} offers
an alternative that explains the ordering of items. Assume that there
exists latent utilities $\left\{ u_{i}\right\} $, one per item $\left\{ \pi_{i}\right\} $.
The ordering $P^{*}(\pib)$ is defined as $\Pr\left(u_{1}\ge u_{2}\ge...\ge u_{N}\right)$.
Here we show that it is linked to \emph{Gompertz distribution}. Let
$u_{j}\ge0$ denote the latent random utility of item $\pi_{j}$.
Let $v_{j}=e^{bu_{j}}$, the Gompertz distribution has the PDF $P_{j}(u_{j})=b\eta_{j}v_{j}\exp\left(-\eta_{j}v_{j}+\eta_{j}\right)$
and the CDF $F_{j}(u_{j})=1-\exp\left(-\eta_{j}(v_{j}-1)\right)$,
where $b>0$ is the scale and $\eta>0$ is the shape parameter. 

At rank $i$, choosing the worst item $\pi_{i}$ translates to ensuring
$u_{i}\le u_{j}$ for all $j<i$. The random utility theory states
that probability of choosing $\pi_{i}$ can be obtained by integrating
out all latent utilities subject to the inequality constraints: 

\begin{eqnarray*}
Q(\pi_{i}\mid\boldsymbol{\pi}_{i+1:N}) & = & \int_{0}^{+\infty}P_{i}(u_{i})\left[\int_{u_{i}}^{+\infty}\prod_{j<i}P_{j}(u_{j})du_{j}\right]du_{i}\\
 & = & \int_{0}^{+\infty}P_{i}(u_{i})\prod_{j<i}\left(1-F_{j}(u_{i})\right)du_{i}\\
 & = & \int_{0}^{+\infty}b_{i}\eta_{ii}v_{i}\exp\left(-v_{i}\sum_{j\le i}\eta_{ij}+\sum_{j\le i}\eta_{ij}\right)du_{i}
\end{eqnarray*}
Note that we have used $\eta_{ij}$ instead of $\eta_{j}$ since this
location parameter is specific to rank $i$. Defining $\eta_{ij}=\exp\left(-f(\xb_{\pi_{j}})\right)$
and changing the variable from $u_{i}$ to $v_{i}$ gives us: 
\begin{align*}
Q(\pi_{i}\mid\boldsymbol{\pi}_{i+1:N}) & =\int_{1}^{+\infty}\eta_{ii}\exp\left(-v_{i}\sum_{j\le i}\eta_{ij}+\sum_{j\le i}\eta_{ij}\right)dv_{i}\\
 & =\frac{\eta_{ii}e^{\sum_{j\le i}\eta_{j}}}{\sum_{j\le i}\eta_{ij}}\int_{1}^{+\infty}\exp\left(-v_{i}\sum_{j\le i}\eta_{ij}\right)dv_{i}\\
 & =\frac{\eta_{ii}}{\sum_{j\le i}\eta_{ij}}=\frac{\exp\left(-f(\xb_{\pi_{i}})\right)}{\sum_{j=1}^{i}\exp\left(-f(\xb_{\pi_{j}})\right)}.
\end{align*}
This resembles Eq.~(\ref{eq:backward-local-prob}).

\paragraph{Remark:}

We note in passing that in case of plain Plackett-Luce, $P(\pib)\equiv P^{*}(\pib)$
if and only if $u_{\pi_{j}}$ is drawn from a Gumbel distribution
of location $f(\xb_{\pi_{j}})$ and scale $1$ for all $j=1,2,...,N$
\cite{mcfadden1973conditional,yellott1977relationship}.

\subsection{Linear time learning}

We show that both the loss and its functional gradient can be computed
in linear time. Let $\phi_{i}=\exp\left(-f(\xb_{\pi_{i}})\right)$,
$Z_{i}=\sum_{j=1}^{i}\phi_{j}=\phi_{i}+Z_{i-1}$ , and $Y_{i}=\sum_{j=1}^{i}1/Z_{j}$.
The loss in Eq.~(\ref{eq:rev-PL-loss}) becomes $\ell_{2}(\perm)=\sum_{i=1}^{N}\left(f(\xb_{\pi_{i}})+\log Z_{i}\right)$
which can be evaluated in linear time. The functional gradient is
reduced to $\partial_{f_{k}}\ell_{2}=\gamma_{k}\left(1-\phi_{k}Y_{k}\right)$,
which is constant for each $k$.

\section{Neural Highway Networks for Rank Function}

Once the rank models have been specified, it remains to define the
rank function $f(\xb)$. We propose the novel use of neural highway
networks \cite{srivastava2015training} to approximate $f(\xb)$,
under the new ranking loss functions proposed in Eqs.~(\ref{eq:PL-loss},\ref{eq:rev-PL-loss}).
Highway networks are powerful, compact function approximator that
overcomes the major bottleneck of standard deep neural networks: with
increasing depth, it is much harder to pass information and gradient
between input and output. More specifically, highway networks are
a special neural network of $L$ nonlinear projection layers, of which
$L-1$ layers are recurrent. At the bottom, the data is projected
into a $K$-dimensional space $\zb=g\left(\bb_{H}+W_{X}\xb\right)$,
where $g$ is an element-wise non-linear transform. Subsequent layers
are recursive same-dimensional projections:

\begin{align*}
\zb_{l+1} & \leftarrow H(\zb_{l})\ast T(\zb_{l})+\zb_{l}\ast\left(1-T(\zb_{l})\right)
\end{align*}
where $H(\zb)=g\left(\bb_{H}+W_{H}\zb\right)$, $T(\zb)\in\left[0,1\right]^{K}$
is an element-wise gating function, and $\ast$ denotes element-wise
product. The gate $T(\zb)$ allows information to pass freely when
$T(\zb)=0$ and forces total nonlinear transforms when $T(\zb)=1$.
A typical implementation is $T(\zb)=\sigma\left(\bb_{T}+W_{T}\zb\right)$,
where $\sigma$ is the sigmoid function. At the top layer, we have
$f(\xb)=\left\langle \wb,\zb_{L}\right\rangle $. Note that when the
number of hidden layer is 1, this returns to the standard shallow
neural network; and when $T(\zb)=1$, the highway network becomes
a recurrent network. Let $\thetab=\left(\bb_{T},\bb_{H},W_{T},W_{H},W_{X},\wb\right)$
be model parameters, gradient-based learning proceeds as follows:
\[
\thetab\leftarrow\thetab-\frac{\mu}{N}\sum_{k=1}^{N}\frac{\partial\ell_{1,2}}{\partial f(\xb_{\pi_{k}})}\frac{\partial f(\xb_{\pi_{k}})}{\partial\thetab}
\]
for learning rate $\mu>0$, where $\frac{\partial\ell_{1,2}}{\partial f(\xb_{\pi_{k}})}$
are functional gradients derived from Eqs.~(\ref{eq:PL-loss},\ref{eq:rev-PL-loss}).

\section{Experimental Evaluation }

\subsection{Data \& Evaluation Metrics}

We validate the proposed model using a large-scale Web dataset from
the Yahoo! learning to rank challenge \cite{chapelle2011yahoo}, .
The dataset is split into a training set of $18,425$ queries ($425,821$
documents), and a testing set of $1,520$ queries ($47,313$ documents).
We also prepare a smaller training subset, called Yahoo!-small, which
has $1,568$ queries and $47,314$ documents. The Yahoo! datasets
contain the groundtruth relevance scores (from $0$ for irrelevant
to $4$ for perfectly relevant). There are $519$ pre-computed unique
features for each query-document pair. We normalize the features across
the whole training set to have mean $0$ and standard deviation $1$. 

Two evaluation metrics are employed. The Normalized Discount Cumulative
Gain \cite{jarvelin2002cumulated} is defined as:
\[
NG@T=\frac{1}{N_{max}}\sum_{i=1}^{T}\frac{2^{r_{\pi_{i}}}-1}{\log_{2}(1+i)}
\]
where $r_{\pi_{i}}$ is the relevance of item at rank $i$. The other
metric is Expected Reciprocal Rank (ERR) \cite{chapelle2009expected},
which was used in the Yahoo! learning-to-rank challenge (2011): 
\begin{align*}
ERR & =\sum_{i}\frac{R(r_{\pi_{i}})}{i}\prod_{j>i}\left[1-R(r_{\pi_{j}})\right],\,\mbox{s.t.}\,R(r_{\pi_{i}})=\frac{2^{r_{\pi_{i}}}-1}{16}
\end{align*}
Both metrics discount for the long list and place more emphasis on
the top ranking. Finally, all metrics are averaged across all test
queries.

\subsection{Model Implementation}

\begin{figure}
\begin{centering}
\includegraphics[width=0.7\textwidth]{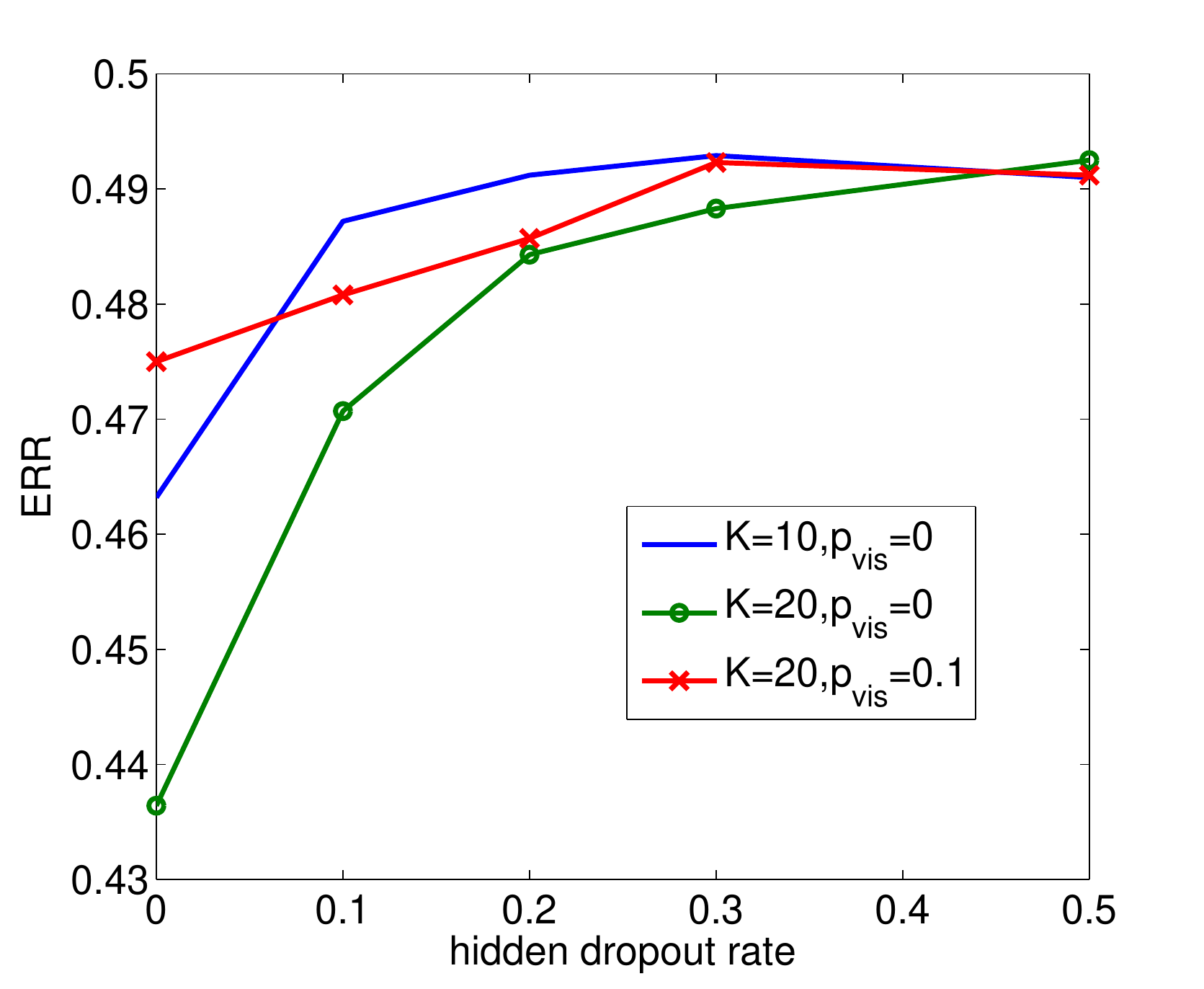}
\par\end{centering}

\protect\caption{Effect of dropouts on NCE performance of a $3$-layer highway nets
on Yahoo! small set. $K$ is number of hidden units, $p_{vis}$ is
dropout rate for visible layer.\label{fig:Effect-of-dropouts}}
\end{figure}

The highway nets are configured as follows. Unless stated otherwise,
we use ReLU units for transformation $H(\zb)$. Parameters $W_{1},W_{H}$
and $W_{T}$ are initialized randomly from a small Gaussian. Gate
bias $\bb_{T}$ is initialized at $-1$ to encourage passing more
information, as suggested in \cite{srivastava2015training}. Transform
bias $\bb_{H}$ is initialized at $0$. To prevent overfitting, dropout
\cite{srivastava2014dropout} is used, i.e., during training for each
item in a mini-batch, input features and hidden units are randomly
dropped with probabilities $p_{vis}$ and $p_{hid}$, respectively.
As dropouts may cause big jumps in gradient and parameters, we set
max-norm per hidden unit to $1$. The mini-batch size is 2 queries,
the learning rate starts from $0.1$, and is halved when there is
no sign of improvement in training loss. Learning stops when learning
rate falls below $10^{-4}$. We fix the dropout rates as follows:
(a) for small Yahoo! data, $p_{vis}=0$, $p_{hid}=0.3$ and $K=10$;
(b) for large Yahoo! data, $p_{vis}=0$, $p_{hid}=0.2$ and $K=20$.
Fig.~\ref{fig:Effect-of-dropouts} shows the effect of dropouts on
the NCE model on the small Yahoo! dataset.

\subsection{Results}

\subsubsection{Performance of choice models.}

\begin{table*}
\begin{centering}
\begin{tabular}{l|ccc|ccc}
\cline{2-7} 
 & \multicolumn{3}{c|}{Yahoo!-small} & \multicolumn{3}{c}{Yahoo!-large}\tabularnewline
\cline{2-7} 
Rank model & ERR & NDCG@1 & NDCG@5 & ERR & NDCG@1 & NDCG@5\tabularnewline
\hline 
Rank SVM & 0.477 & 0.657 & 0.642 & 0.488 & 0.681 & 0.666\tabularnewline
Plackett-Luce & 0.489 & 0.683 & 0.652 & 0.495 & 0.699 & 0.671\tabularnewline
\hline 
Choice by elimination & \textbf{0.497} & \textbf{0.697} & \textbf{0.664} & \textbf{0.503} & \textbf{0.709} & \textbf{0.680}\tabularnewline
\hline 
\end{tabular}
\par\end{centering}

\protect\caption{Performance with linear rank functions. \label{tab:Sequential-models}}
\end{table*}

In this experiment we ask whether the new choice by elimination model
has any advantage over existing state-of-the-arts (the forward selection
method of Plackett-Luce \cite{xia2008listwise_FULL} and the popular
Rank SVM \cite{joachims2002optimizing_FULL}). Table~\ref{tab:Sequential-models}
reports results on the all datasets (Yahoo!-small, Yahoo!-large) on
different sequential models. The NCE works better than Rank SVM and
Plackett-Luce. Note that due to the large size, the differences are
statistically significant. In fact, in the Yahoo! L2R challenge, the
top 20 scores (out of more than 1,500 teams) differ only by 1.56\%
in ERR, which is less than the difference between Plackett-Luce and
choice by elimination (1.62\%).

\subsubsection{Neural nets versus gradient boosting.}

\begin{figure}
\begin{centering}
\begin{tabular}{l|ccc|ccc}
 & \multicolumn{3}{c|}{Placket-Luce} & \multicolumn{3}{c}{Choice by elimination}\tabularnewline
\hline 
Rank function & ERR & NDCG@1 & NDCG@5 & ERR & NDCG@1 & NDCG@5\tabularnewline
\hline 
SGTB & 0.497 & 0.697 & 0.673 & 0.506 & 0.705 & 0.681\tabularnewline
Neural nets & \textbf{0.501} & \textbf{0.705} & \textbf{0.688} & \textbf{0.509} & \textbf{0.719} & \textbf{0.697}\tabularnewline
\hline 
\end{tabular}
\par\end{centering}

\protect\caption{Comparing highway neural networks against stochastic gradient tree
boosting (SGTB) on Yahoo!-large dataset. \label{tab:Comparing-deep-vs-tree}}

\end{figure}

We also compare the highway networks against the best performing method
in the Yahoo! L2R challenge. Since it was consistently demonstrated
that gradient boosting trees work best \cite{chapelle2011yahoo},
we implement a sophisticated variant of Stochastic Gradient Tree Boosting
(SGTB) \cite{Friedman-CSDA02} for comparison. In SGTB, at each iteration,
regression trees are grown to fit a random subset of functional gradients
$\frac{\partial\ell}{\partial f(\xb)}$. Grown trees are added to
the ensemble with an adaptive learning rate which is halved whenever
the loss fluctuates and does not decrease. At each tree node, a random
subset of features is used to split the node, following \cite{breiman2001random}.
Tree nodes are split at random as it leads to much faster tree growing
without hurting performance \cite{geurts2006extremely}. The SGTB
is configured as follows: number of trees is \textbf{$300$}; learning
rate starts at $0.1$; a random subset of $50\%$ data is used to
grow a tree; one third of features are randomly selected at each node;
trees are grown until either the number of leaves reaches $512$ or
the node size is below $40$.

Table.~\ref{tab:Comparing-deep-vs-tree} show performance scores
of models trained under different losses on the Yahoo!-large dataset.
The highway neural networks are consistently competitive against the
gradient tree boosting, the best performing rank function approximator
in this challenge \cite{chapelle2011yahoo}.

\section{Conclusion\label{sec:Conclusion}}

We have presented \emph{Neural Choice by Elimination}, a new framework
that integrates deep neural networks into a formal modeling of human
behaviors in making sequential choices. Contrary to the standard Plackett-Luce
model, where the most worthy items are iteratively selected, here
the least worthy items are iteratively eliminated. Theoretically we
show that choice by elimination is equivalent to sequentially marginalizing
out Thurstonian random utilities that follow Gompertz distributions.
Experimentally we establish that deep neural networks are competitive
in the learning to rank domain, as demonstrated on a large-scale public
dataset from the Yahoo! learning to rank challenge. 

\bibliographystyle{plain}

\begin{thebibliography}{10}

\bibitem{agarwal2012learning}
Arvind Agarwal, Hema Raghavan, Karthik Subbian, Prem Melville, Richard~D
  Lawrence, David~C Gondek, and James Fan.
\newblock Learning to rank for robust question answering.
\newblock In {\em Proceedings of the 21st ACM international conference on
  Information and knowledge management}, pages 833--842. ACM, 2012.

\bibitem{azari2012random}
Hossein Azari, David Parks, and Lirong Xia.
\newblock Random utility theory for social choice.
\newblock In {\em Advances in Neural Information Processing Systems}, pages
  126--134, 2012.

\bibitem{bengio2009learning}
Yoshua Bengio.
\newblock {Learning deep architectures for AI}.
\newblock {\em Foundations and trends{\textregistered} in Machine Learning},
  2(1):1--127, 2009.

\bibitem{breiman2001random}
L.~Breiman.
\newblock Random forests.
\newblock {\em Machine learning}, 45(1):5--32, 2001.

\bibitem{breiman96bagging}
Leo Breiman.
\newblock Bagging predictors.
\newblock {\em Machine Learning}, 24(2):123--140, 1996.

\bibitem{burges2005learning_FULL}
C.~Burges, T.~Shaked, E.~Renshaw, A.~Lazier, M.~Deeds, N.~Hamilton, and
  G.~Hullender.
\newblock {Learning to rank using gradient descent}.
\newblock In {\em Proceedings of the 22nd international conference on Machine
  learning}, page~96. ACM, 2005.

\bibitem{burges2011learning}
Christopher~JC Burges, Krysta~Marie Svore, Paul~N Bennett, Andrzej Pastusiak,
  and Qiang Wu.
\newblock Learning to rank using an ensemble of lambda-gradient models.
\newblock In {\em Yahoo! Learning to Rank Challenge}, pages 25--35, 2011.

\bibitem{chapelle2011yahoo}
O.~Chapelle and Y.~Chang.
\newblock Yahoo! learning to rank challenge overview.
\newblock In {\em JMLR Workshop and Conference Proceedings}, volume~14, pages
  1--24, 2011.

\bibitem{chapelle2009expected}
O.~Chapelle, D.~Metlzer, Y.~Zhang, and P.~Grinspan.
\newblock {Expected reciprocal rank for graded relevance}.
\newblock In {\em CIKM}, pages 621--630. ACM, 2009.

\bibitem{deng2013deep}
Li~Deng, Xiaodong He, and Jianfeng Gao.
\newblock Deep stacking networks for information retrieval.
\newblock In {\em Acoustics, Speech and Signal Processing (ICASSP), 2013 IEEE
  International Conference on}, pages 3153--3157. IEEE, 2013.

\bibitem{dong2014rankcnn}
Yuan Dong, Chong Huang, and Wei Liu.
\newblock Rankcnn: When learning to rank encounters the pseudo preference
  feedback.
\newblock {\em Computer Standards \& Interfaces}, 36(3):554--562, 2014.

\bibitem{Friedman-ann-stat01}
Jerome~H. Friedman.
\newblock Greedy function approximation: a gradient boosting machine.
\newblock {\em Annals of Statistics}, 29(5):1189--1232, 2001.

\bibitem{Friedman-CSDA02}
J.H. Friedman.
\newblock Stochastic gradient boosting.
\newblock {\em Computational Statistics and Data Analysis}, 38(4):367--378,
  2002.

\bibitem{fu2014reverse}
Qiang Fu, Jingfeng Lu, and Zhewei Wang.
\newblock 'reverse' nested lottery contests.
\newblock {\em Journal of Mathematical Economics}, 50:128--140, 2014.

\bibitem{geurts2006extremely}
Pierre Geurts, Damien Ernst, and Louis Wehenkel.
\newblock Extremely randomized trees.
\newblock {\em Machine learning}, 63(1):3--42, 2006.

\bibitem{Gormley-Murphy08}
Isobel~Claire Gormley and Thomas~Brendan Murphy.
\newblock A mixture of experts model for rank data with applications in
  election studies.
\newblock {\em The Annals of Applied Statistics}, 2(4):1452--1477, 2008.

\bibitem{henery1981permutation}
RJ~Henery.
\newblock Permutation probabilities as models for horse races.
\newblock {\em Journal of the Royal Statistical Society, Series B},
  43(1):86--91, 1981.

\bibitem{huang2013learning}
Po-Sen Huang, Xiaodong He, Jianfeng Gao, Li~Deng, Alex Acero, and Larry Heck.
\newblock Learning deep structured semantic models for web search using
  clickthrough data.
\newblock In {\em Proceedings of the 22nd ACM international conference on
  Conference on information \& knowledge management}, pages 2333--2338. ACM,
  2013.

\bibitem{jarvelin2002cumulated}
K.~J{\"a}rvelin and J.~Kek{\"a}l{\"a}inen.
\newblock {Cumulated gain-based evaluation of IR techniques}.
\newblock {\em ACM Transactions on Information Systems (TOIS)}, 20(4):446,
  2002.

\bibitem{joachims2002optimizing_FULL}
T.~Joachims.
\newblock {Optimizing search engines using clickthrough data}.
\newblock In {\em Proceedings of the eighth ACM SIGKDD international conference
  on Knowledge discovery and data mining}, pages 133--142. ACM New York, NY,
  USA, 2002.

\bibitem{li2007mcrank}
P.~Li, C.~Burges, Q.~Wu, JC~Platt, D.~Koller, Y.~Singer, and S.~Roweis.
\newblock Mcrank: Learning to rank using multiple classification and gradient
  boosting.
\newblock {\em Advances in neural information processing systems}, 2007.

\bibitem{liu2011learning}
Tie-Yan Liu.
\newblock {\em Learning to rank for information retrieval}.
\newblock springer, 2011.

\bibitem{mcfadden1973conditional}
D.~McFadden.
\newblock Conditional logit analysis of qualitative choice behavior.
\newblock {\em Frontiers in Econometrics}, pages 105--142, 1973.

\bibitem{plackett1975analysis}
R.L. Plackett.
\newblock {The analysis of permutations}.
\newblock {\em Applied Statistics}, pages 193--202, 1975.

\bibitem{severyn2015learning}
Aliaksei Severyn and Alessandro Moschitti.
\newblock Learning to rank short text pairs with convolutional deep neural
  networks.
\newblock In {\em Proceedings of the 38th International ACM SIGIR Conference on
  Research and Development in Information Retrieval}, pages 373--382. ACM,
  2015.

\bibitem{song2014adapting}
Yang Song, Hongning Wang, and Xiaodong He.
\newblock Adapting deep ranknet for personalized search.
\newblock In {\em Proceedings of the 7th ACM international conference on Web
  search and data mining}, pages 83--92. ACM, 2014.

\bibitem{srivastava2014dropout}
Nitish Srivastava, Geoffrey Hinton, Alex Krizhevsky, Ilya Sutskever, and Ruslan
  Salakhutdinov.
\newblock Dropout: A simple way to prevent neural networks from overfitting.
\newblock {\em Journal of Machine Learning Research}, 15:1929--1958, 2014.

\bibitem{srivastava2015training}
Rupesh~Kumar Srivastava, Klaus Greff, and J{\"u}rgen Schmidhuber.
\newblock Training very deep networks.
\newblock {\em arXiv preprint arXiv:1507.06228}, 2015.

\bibitem{thurstone1927law}
L.L. Thurstone.
\newblock A law of comparative judgment.
\newblock {\em Psychological review}, 34(4):273, 1927.

\bibitem{truyen_phung_venkatesh_icml13}
T.~Tran, D.~Phung, and S.~Venkatesh.
\newblock {Thurstonian Boltzmann Machines: Learning from Multiple
  Inequalities}.
\newblock In {\em International Conference on Machine Learning (ICML)},
  Atlanta, USA, June 16-21 2013.

\bibitem{Truyen:2012b}
T.~Tran, D.Q. Phung, and S.~Venkatesh.
\newblock Sequential decision approach to ordinal preferences in recommender
  systems.
\newblock In {\em Proc. of the 26th AAAI Conference}, Toronto, Ontario, Canada,
  2012.

\bibitem{Truyen:2011a}
T.~Truyen, D.Q Phung, and S.~Venkatesh.
\newblock Probabilistic models over ordered partitions with applications in
  document ranking and collaborative filtering.
\newblock In {\em Proc. of SIAM Conference on Data Mining (SDM)}, Mesa,
  Arizona, USA, 2011. SIAM.

\bibitem{tversky1972elimination}
Amos Tversky.
\newblock Elimination by aspects: A theory of choice.
\newblock {\em Psychological review}, 79(4):281, 1972.

\bibitem{xia2008listwise_FULL}
F.~Xia, T.Y. Liu, J.~Wang, W.~Zhang, and H.~Li.
\newblock {Listwise approach to learning to rank: theory and algorithm}.
\newblock In {\em Proceedings of the 25th international conference on Machine
  learning}, pages 1192--1199. ACM, 2008.

\bibitem{yellott1977relationship}
John~I Yellott.
\newblock {The relationship between Luce's choice axiom, Thurstone's theory of
  comparative judgment, and the double exponential distribution}.
\newblock {\em Journal of Mathematical Psychology}, 15(2):109--144, 1977.

\end{thebibliography}

\end{document}